\pdfoutput=1
\documentclass{article}

\usepackage[final]{nips_2018}
\usepackage{amsmath} 
\usepackage{amsmath}
\usepackage{amssymb}
\usepackage{bbm}
\usepackage{verbatim}
\usepackage{tkz-graph}
\usetikzlibrary{arrows}
\usepackage{caption}
\usepackage{subcaption}
\usepackage[11pt]{moresize}

\DeclareMathOperator{\E}{\mathbb{E}}

\newcommand*{\rom}[1]{\expandafter\@slowromancap\romannumeral #1@}
\makeatother
\renewcommand{\P}{\mathbb{P}}

\newcommand{\Real}{\mathbb{R}}

\newcommand{\cemr}{\operatorname{cemr}}
\newcommand{\osr}{\operatorname{osr}}

\usepackage{natbib}
\newcommand{\RNum}[1]{\uppercase\expandafter{\romannumeral #1\relax}}

\newcommand{\states}{\mathcal{S}}
\newcommand{\actions}{\mathcal{A}}

\newcommand{\statecount}{S}
\newcommand{\actioncount}{A}

\begin{document}
	\author{Reazul Hasan Russel\\
		Department of Computer Science\\
		University of New Hampshire\\
		Durham, NH-03824 USA\\
		{\tt rrussel@cs.unh.edu}}
	\title{A Short Survey on Probabilistic Reinforcement Learning}
	
	\maketitle{} 
	
	
	\begin{abstract}
	A reinforcement learning agent tries to maximize its cumulative payoff by interacting in an unknown environment. It is important for the agent to explore suboptimal actions as well as to pick actions with highest known rewards. Yet, in sensitive domains, collecting more data with exploration is not always possible, but it is important to find a policy with a certain performance guaranty. In this paper, we present a brief survey of methods available in the literature for balancing exploration-exploitation trade off and computing robust solutions from fixed samples in reinforcement learning.
	\end{abstract}
	
	\section{Introduction}
	Reinforcement Learning is learning to map situations to actions that maximize a long term objective~\citep{Sutton1998, Puterman2005, Csaba2010}. The actions are not labeled for training, rather the agent needs to learn about most rewarding actions by trying them. An action affects both the immediate reward and the next state yielding short and long term consequences. An important element of some reinforcement learning systems is a model of the environment, which mimics its behavior.
	
	Markov Decision Process (MDP) is a mathematical framework to build the model of the environment for reinforcement learning~\citep{Bertsekas1996,Sutton1998,Puterman2005}. An MDP is a tuple, $\mathcal{M} = (\states, \actions, R, P, H, p_0)$, where $\states=\{1, \ldots, S\}$ is the state space, $\actions=\{1,\ldots,A\}$ is the action space, $H$ is the horizon and $p_0$ is the initial state distribution. At each time period $h=1,\ldots,H$ within an episode, the agent observes a state $s_h \in \states$, takes action $a_h \in \actions$, receives a reward $r_h \in R(s_h, a_h)$ and transitions to a new state $s_{h+1} \sim P(s_h, a_h)$. A \emph{reward signal} $r\in R$ is a scalar defining the desirability of an event, which the agent wants to maximize over long run. A policy $\pi$ is a mapping from a state $s \in \states$ to an action $a \in \actions$. A \emph{value function} $V_{\pi}(s)$ provides an estimate of the expected amount of total reward the agent can accumulate over the future when following a policy $\pi$, starting from any particular state $s$. The return $\rho(s,\pi)$ of an MDP is the total accumulated reward for executing a policy $\pi$ starting from a state $s$. 
	
	MDPs usually assume that transition probabilities are known precisely, but this is rarely the case in reality. Corrupted data set, noisy data collection process, missing assumptions, overlooked factors etc. are common reasons for errors in transition probabilities~\citep{Petrik2014,Wiesemann2013,Iyengar2005}. Also, a finite set of samples may not exactly represent the underlying true transition model. This uncertainty often leads to policies that fail in real-world deployments. 
	
	One of the main challenges in learning the transition probabilities is the trade-off between exploration and exploitation~\citep{Sutton1998}. The agent needs to prefer actions yielding higher rewards, but also needs to discover such actions. Different actions need to be considered to collect statistically significant evidence and then favor those with higher rewards. \emph{Optimism in the face of uncertainty} promotes exploration by encouraging the agent to try less certain but high potential actions~\citep{Brafman2001, Auer2006, Dann2015, Cserna2017, Osband2017}.
	
	On the other hand, in a \emph{batch RL} setup, a fixed training set based on the historic interaction with the environment is provided. More data cannot be collected at will in this situation because of the sensitivity of high stake domains, but a solution with certain performance guaranty is still important to obtain. For example, it can reduce the chance of an unpleasant surprise when the policy is deployed and can be used to justify the need to collect more data~\citep{Petrik2016a,Lim2013,Hanasusanto2013}. If the lower bound on the return is smaller than the return of the currently-deployed policy, then the currently deployed policy would not be replaced.
	
	In this paper, we review different approaches proposed in the literature for dealing with uncertainty in reinforcement learning. We analyze both the optimistic and robust formulations of the problem. We discuss different proposed methods and their advantages/disadvantages. The paper is organized as follows: in Section \ref{sec:exploration}, we present different methods proposed for exploration-exploitation balancing. Section \ref{sec:robustness} discusses methods available for computing robust policies in batch RL setup. Finally, we draw the concluding remarks in Section \ref{sec:conclusion} 
	
	\section{Exploration} \label{sec:exploration}
	A reinforcement learning agent faces a fundamental trade-off between exploration and exploitation. The question of exploration is \emph{"how can learning time be minimized?"} and the question of exploitation is \emph{"how can rewards be maximized?"}. Pure exploration does not minimize learning time, because the agent may attempt sub-optimal actions. Similarly, pure exploitation cannot maximize the reward, because the agent may keep choosing a sub-optimal action~\citep{Thrun1992}. A reinforcement learning agent tries to find an optimal policy while minimizing both learning time and cost, which boils down to balancing exploration and exploitation. This is a very active research area in reinforcement learning and has received significant attention.
	
	\subsection{Directed/Undirected Exploration}
	\cite{Thrun1992b} states two broad categories of exploration techniques: 1) Undirected exploration, and 2) Directed exploration.
	
	\textbf{Undirected exploration} Undirected exploration always picks an action randomly without considering the previous experience or the underlying learning process: $\P(a) = \frac{1}{\actioncount}$. This doesn't learn anything and is the worst possible exploration strategy in terms of cost. A simple alternative is to explore uniformly with a small probability $\epsilon$, and act greedily rest of the time~\citep{Sutton1998, Tokic2010}.
	
	\[ \P(a) = 
	\left\{
	\begin{array}{ll}
	1-\epsilon, & \text{if $a$ maximizes return}\\
	\frac{1}{\text{\actioncount}}, & \text{otherwise} \\
	\end{array} 
	\right.
	\]
	
	This method is known as $\epsilon$-$greedy$. In the limit, as the number of steps increases, $\epsilon$-$greedy$ converges with the guarantee that each action is sampled an infinite number of times~\citep{Sutton2016}. The $\epsilon$-$greedy$ method is simple and effective, but it keeps choosing a sub-optimal action with small probability in the long run. Decaying $\epsilon$ over time can minimize this sub-optimal exploration, but it is still non-trivial to set this decay-rate  reasonably due to its non-adaptive nature.
	
	\textbf{Directed exploration} Directed exploration considers the underlying learning process and utilizes the learned knowledge to guide exploration. A basic directed exploration strategy is counter based exploration where the number of visits to each state and/or action is maintained. A policy is evaluated with exploitation value and exploration bonus based on collected statistics. \emph{Dyna} algorithm~\citep{Sutton1991} maintains visitation statistics for each transition from state $s$ to $s'$ with action $a$ and obtaining reward $r$. It updates the policy at a state $s$ with the latest model as:
	
	\begin{equation} \label{eq:Bellman}
		Q(s,a) = \hat{R}(s,a) + \gamma\sum_{s'}\hat{P}(s,a,s')\max_{a'}Q(s',a')
	\end{equation}
	
	Here, $Q(s,a)$ is the state-action value for taking an action $a$ in a state $s$ and $\gamma$ is the discount factor. Dyna performs $k$ additional random updates following the same update model. This adds more computational cost for Dyna than general Q-learning~\citep{Watkins1992}, but Dyna requires an order of magnitude fewer steps to converge to an optimal policy.
	
	\cite{Kocsis2006} proposes a roll-out based Monte-Carlo planning algorithm \emph{UCT} (Upper Confidence Bound applied to trees). UCT incrementally builds a lookahead tree by repeatedly sampling state-action-reward triplets and returns the action with highest average observed reward. \cite{Gelly2007} proposes three different variations of UCT and applies them in \emph{MoGo} Go program. A simple count-based technique for deep-RL is described in \cite{Tang2017}; they map states with a hash function and derive an exploration bonus using count statistics. 
	
	However, it is not a surprise that directed exploration techniques usually outperform undirected exploration techniques in most problem setups.
	
	\subsection{Exploration with Different Objectives}
	Another main consideration in classifying exploration techniques is the underlying objective of the optimization problem. Researchers usually consider three different objectives to optimize:
	
	\begin{enumerate}
		\item Probably Approximately Correct-MDPs (PAC-MDPs)
		\item Knows What It Knows (KWIK)
		\item Regret Bounds
	\end{enumerate}
	
	\textbf{PAC-MDPs} For optimal exploration, an RL algorithm needs to obtain the maximum possible expected discounted reward $\E[\sum_{t=1}^{\infty}\gamma^{t-1}r_t]$. This is a very hard problem and only tractable for a special class of MDPs known as K-armed bandits~\citep{Gittins1979}. A relaxed but still challenging problem is to act near-optimally in terms of two additional positive real parameters $\epsilon$ and $\delta$~\citep{Kakade2003, Strehl2008a}. Here $\epsilon$ defines the quality of the solution obtained by the algorithm and $\delta$ is a measure of confidence on the obtained solution. For any $\epsilon>0$ and $0<\delta<1$, the complexity of an efficient PAC-MDP algorithm is bounded by polynomials in terms of $(S, A, \frac{1}{\epsilon},\frac{1}{\delta},\frac{1}{1-\gamma})$ with probability at least $1-\delta$~\citep{Strehl2008a,Strehl2009}. One important notion in analyzing PAC-optimal algorithms is the \emph{$\epsilon$-return mixing time $T$} for a policy $\pi$, which is the smallest number $T$ of steps required to guarantee that the distribution of states after t steps of $\pi$ is within $\epsilon$ of the stationary distribution induced by $\pi$ \citep{Kearns1998a}.
	
	A model-based algorithm \emph{Explicit Explore or Exploit $(E^3)$} proposed by \cite{Kearns1998a} estimates the MDP parameters from data and finds a policy based on the estimated parameters. $E^3$ divides the states into two sets: known and unknown. States those are visited at least $m$ times (for some number $m$) having an accurate estimate of transition probabilities and rewards with high probability are labeled as known. $E^3$ maintains two estimated MDPs, $\hat{M}$ and $\hat{M'}$ based on the estimated parameters for known states. In $\hat{M'}$, the unknown states are merged into a recurrent state with maximum reward $R_{max}$, encouraging exploration. In an unknown state, $E^3$ picks the action that was chosen the fewest number of times (balanced wandering). When a known state appears during balanced wandering, $E^3$ exploits if the exploitation policy $\hat{\pi}$ for $\hat{M}$ achieves a return that is at most $\frac{\epsilon}{2}$ away from the true return. Otherwise, it executes the exploration policy $\hat{\pi'}$ derived from $\hat{M'}$. $E^3$ returns a policy that is within $\epsilon$ of the true policy in terms of return with probability at least $1-\delta$ and complexity polynomial in $(\frac{1}{\epsilon}, \frac{1}{\delta}, \statecount, \frac{1}{1-\alpha}, R_{max})$.
	
	\cite{Brafman2001} proposes \emph{R-{\ssmall MAX}}, a model-based RL algorithm with near-optimal average reward in polynomial time. R-{\ssmall MAX} initializes the model optimistically by assigning maximum possible rewards for each state-action. It maintains an empirical maximum-likelihood estimate for reward and transition distributions of each state-action pair and updates them based on the agent's observation. Action-selection step always picks the action that maximizes the state-action value $Q(s,.)$ and the update step solves the  Bellman equation in \ref{eq:Bellman}. R-{\ssmall MAX} achieves an expected return within $2\epsilon$ of true return with probability at least $1-\delta$ and complexity polynomial in $(N, k, T, \frac{1}{\epsilon}, \frac{1}{\delta})$. Here, $N$ is the number of states, $k$ is the number of actions, $T$ is the $\epsilon$-return mixing time of an optimal policy.
	
	\textbf{Knows What It Knows (KWIK)} \cite{Li2011} proposes the KWIK learning framework of supervised learning for problems where active exploration affects the generated training samples. The main motivation of KWIK is that, a sample that is independent from the available samples can contribute more in the learning process and can be a good candidate for exploration. In KWIK setup, the algorithm gives a prediction when its accuracy is not more than $\epsilon$ away from the true label. Otherwise, the learning algorithm commits \emph{"don't know"} and requests the true label to update the model. KWIK algorithms do not bound the number of mistakes for performance measure, rather it bounds the number of label requests it can make. This independence of performance measure from the underlying distribution makes KWIK algorithms lucrative for problem domains where the input distribution is dependent in complex ways. \cite{Li2011} presents a different class of algorithms and examples within KWIK framework for deterministic and probabilistic observations along with methods for composing complex hypothesis classes.
	
	\textbf{Regret Bounds} Algorithms considering regret bounds as an objective usually care about the online performance of the algorithm over some finite episodes $K$. For a specific episode $k \in K$ and a policy $\pi$, the regret with respect to the true MDP $M^*$ is defined as:
	
	\[
	\Delta_k = \sum_{s \in \states} p_0(s) (V^{M^*}_{\pi^*}(s) - V^{M^*}_{\pi_k}(s))
	\]
	
	The total regret over $T$ episodes is computed as:
	
	\[
	Regret(T,\pi,M^*) = \sum_{k=1}^{T} \Delta_k
	\]
	
	\cite{Auer2006} presents \emph{UCRL2}, an algorithm that considers the total regret and behaves optimistically in the face of uncertainty (OFU). UCRL2 uses samples available so far to compute the maximum likelihood estimates of transition probabilities $\hat{p}$ and rewards $\hat{r}$. It then defines a set of statistically plausible MDPs $\mathcal{M}$ in terms of confidence regions $d$ from $\hat{p}$ and $\hat{r}$. It picks the most optimistic MDP $\tilde{M} \in \mathcal{M}$ and computes the near optimal policy $\tilde{\pi}$ for $\tilde{M}$ using extended value iteration:
	\[
	v_{i+1}(s) = \max_{a \in \actions} \bigg\{ \tilde{r}(s,a) + \max_{p \in \mathcal{P}(s,a)} \Big\{ \sum_{s' \in \states} p(s')v_i(s') \Big\} \bigg\}
	\]
	Here, $\tilde{r}(s,a)$ and $\mathcal{P}(s,a)$ are the maximum plausible estimates of reward and transition probabilities within the confidence region $d$. Computation of $\tilde{\pi}$ happens only at the beginning of the episode and execution of $\tilde{\pi}$ generates samples to be used in the next episode. The regret of UCRL2 is bounded by $34DS\sqrt{AT \log(\frac{T}{\delta})}$ with probability at least $1-\delta$. 
	
	\cite{Strens2002, Osband2013, Osband2016} propose a Bayesian \emph{stochastically optimistic} algorithm \emph{Posterior Sampling Reinforcement Learning (PSRL)} inspired by the idea of Thompson Sampling \citep{Thompson1933}. PSRL proceeds in episodes. At the start of each episode $k$, it updates a prior distribution over MDPs with states $\states$, actions $\actions$ and horizon $\tau$. It then samples a single statistically plausible MDP $\mathcal{M}_k$ from the posterior and selects a policy $\mu_k$ to maximize value for that MDP. PSRL follows this policy $\mu_k$ over episode $k$ and collects samples for the next episode. PSRL is conecptually simple and computationally efficient. The Bayesian nature of the algorithm allows it to incorporate any prior knowledge. OFU algorithms like UCRL2 are often overly conservative because they bound worst-case scenarios simultaneously for every state-action pair \citep{Osband2016, Petrik2014, Wiesemann2013, Osband2013}. But PSRL doesn't need to explicitly construct the statistically complicated confidence bounds. PSRL rather selects policies with proportional probabilities of being optimal, quantifying the uncertainty efficiently with posterior distribution. Therefore, PSRL outperforms OFU algorithms by a significant margin. The Bayesian regret of PSRL algorithm is bounded by $H\sqrt{SAT}$.

	\section{Uncertainty and Robustness} \label{sec:robustness}
	
	Uncertainty is inherent in many real world problems. Often it is important to ensure reasonable performance guarantee during the process of learning or deployment in those problems. Robustness cares about learning a policy that maximizes the expectation of return with respect to worst-case scenario \citep{Ben-Tal2009, Petrik2014}:
	
	\begin{equation} \label{eq:general_robust}
	\max_{\pi \in \Pi} \min_{w \in \Omega^\pi} \E_{\pi,w}\bigg( \sum_{t=0}^{\infty} \gamma^tr_t \bigg)
	\end{equation}
	
	Here, $\Omega^\pi$ is a set of trajectories $(s_0,a_0,s_1,a_1,\ldots)$ under policy $\pi$, $\E_{\pi,w}(\cdot)$ is the expected return under policy $\pi$ and transition probabilities inferred from trajectory $w$. This objective mitigates the effects of stochastic nature of the problem and uncertainty about problem parameters \citep{Nilim2005, Tamar2013, Garc2018}. One popular approach to optimize this objective is to build an approximate model of the environment first and then apply dynamic programming algorithms to learn the optimal policy. Robust MDP (RMDP) is a framework to do just that. RMDPs are identical to normal MDPs as described before, except that the transition probability is not known exactly, rather comes from a set of possible transition matrices $\mathcal{P}(s,a)$ known as an \emph{ambiguity set}~\citep{Iyengar2005, Wiesemann2013, Petrik2014}. Once an action $a\in\actions$ is chosen in state $s\in\states$, a transition probability $p(\cdot|s,a) \in \mathcal{P}(s,a)$ leads to a next state $s' \in \states$ yielding an immediate reward $\hat{r}$. 	The optimal value function of RMDP must satisfy the robust Bellman equation:
	
	\begin{equation} \label{eq:rmdp_obj}
	V(s) = \max_{a \in \actions} \bigg( \hat{r}(s,a) + \min_{p \in \mathcal{P}(s,a)} \Big( \sum_{s' \in \states} p(s')V(s') \Big) \bigg)
	\end{equation}
	
	The quality of the solution on an RMDP depends on the size and shape of the ambiguity set. One important notion about constructing ambiguity sets is the underlying \emph{rectangularity} assumption. Rectangularity simply indicates the independence of the transition probabilities for different states $s\in\states$ or state-action pairs $(s,a)\in\states\times\actions$. the optimization problem in \ref{eq:rmdp_obj} is NP-hard in general, but solvable in polynomial time when $\mathcal{P}$ is $s,a$-rectangular or $s$-rectangular and convex \citep{Nilim2004, Iyengar2005, Wiesemann2013, Ho2018}. In $s,a$-rectangular RMDPs, ambiguity sets are defined independently for each state $s$ and action $a$: $p_{s,a}\in\mathcal{P}_{s,a}$. The most common method for defining ambiguity sets is to use norm-bounded distance from a nominal probability distribution $\bar{p}$ as:
	$$
	\mathcal{P}_{s,a} = \{p \in \Delta^s : \lVert p-\bar{p}_{s,a} \rVert \le \psi_{s,a} \}
	$$
	for a given $\psi_{s,a} \ge 0$, a nominal point $\bar{p}_{s,a}$ and a norm $\lVert\cdot\rVert$. Here, $\Delta^s$ denotes the probability simplex in $\Real_+^S$. The nominal
point is often the most likely transition probability, but needs
not be. In $s$-rectangular RMDPs, the ambiguity set is defined independently for each state $s$: $p_s \in \mathcal{P}_s$ as:
	$$
	\mathcal{P}_s = \Big\{ p_1\in\Delta^s,\ldots,p_A\in\Delta^s : \sum_{a\in\actions} \lVert \bar{p}_{s,a}-p_a \rVert \le \psi_s \Big\}
	$$
	for a given $\psi_s\ge0$ \citep{Wiesemann2013, Ho2018}. 
	
	Based on different optimization objectives, robust RL techniques can be classified into three broad categories: 1) Robust objective 2) Robust Baseline Regret, and 3) Regret based Robust

	\textbf{Robust objective} The general idea of robustness is to compute a policy with the best worst-case performance guarantee. \cite{Nilim2004} considers a robust MDP with uncertain transition probabilities. Their proposed ways to construct ambiguity sets include: log likelihood region based ellipsoidal or interval model, maximum a posteriori model that can incorporate prior information and entropy bound model based on Kullback-Leibler divergence between two distributions. They show that these ambiguity sets are statistically accurate and computationally tractable.
	They prove that perfect duality holds in the robust control problem:
	$\max_{\pi \in \Pi} \min_{p \in \mathcal{P}} V(\pi,p) = \min_{p \in \mathcal{P}} \max_{\pi \in \Pi} V(\pi,p)
	$.
	They solve a robust problem equivalent to equation (\ref{eq:rmdp_obj}) and argues that the computational cost is asymptotically same as of the classical recursion. \cite{Iyengar2005} builds on top of the ideas presented in \cite{Nilim2004} to build ambiguity sets and proposes a unifying framework for robust Dynamic Programming. They present robust value iteration, robust policy iteration algorithms to compute $\epsilon$-optimal robust solutions and independently verifies that the computational cost to solve robust DP is just modestly higher than the non-robust DP.
	
	\cite{Xu2012} proposes a distributionally robust criterion where the optimal policy maximizes the expected total reward under the most plausible adversarial probability distribution. 
	They formulate nested uncertainty sets $\mathcal{C}_s^1\subseteq\mathcal{C}_s^1\subseteq\ldots\mathcal{C}_s^1$ implying $n$ different levels of estimated parameters where $\mathcal{C}^i$s are state-wise independent. The policies then can be ranked based on their performance under most adversarial distributions. They reduce this problem into standard RMDP with a single ambiguity set, show that the optimal policy satisfies a Bellman type equation and can be solved in polynomial time.
	
	\cite{Wiesemann2013} uses the observation history $(s_0,a_0,s_1,a_1,\ldots,s_n,a_n)\in(\states\times\actions)^n$ to construct an ambiguity set that contains the MDP's true transition probabilities with the probability at least $1-\delta$. The worst-case total expected reward under any policy $\pi$ over this ambiguity set then provides a valid lower bound on the expected total reward with the confidence at least $1-\delta$. They develop algorithms to solve the robust policy evaluation and improvement problems with $s,a$-reactangular and $s$-reactangular ambiguity sets in polynomial time. They prove that these algorithms are not tractable for non-rectangular ambiguity sets unless $P=NP$.

	Robust Approximations for Aggregated MDPs (RAAM) \citep{Petrik2014} proposes value function approximation with state aggregation for RL problems with very large state spaces. They consider state importance weights determining the relative importance of approximation errors over states. They describe a linear time algorithm to construct $L_1$ constrained rectangular uncertainty sets. RAAM reduces the robust optimization problem as an RMDP and obtains desired robustness guaranty in the aggregated model by assuming bounded worst-case importance weights.
	
	\textbf{Robust Baseline Regret} Building an accurate dynamics of the model is difficult when the data is limited. It's often the case that a baseline policy $\pi_B$ with certain performance guaranty is deployed on an inaccurate model and it's only possible to replace that policy when a new policy $\pi$ is guaranteed to outperform $\pi_B$. Baseline regret is defined as the difference between these two policies in terms of returns:
	$
	\rho(\pi, P^*) - \rho(\pi_B, P^*)
	$, 
	where $\rho(.)$ is the return and $P^*$ is the true transition probability. Robust regret is the regret obtained in the worst plausible scenario and robust baseline regret optimization technique tries to maximize that:
	
	\[
	\pi_R = \arg \max_{\pi \in \Pi} \min_{\xi\in\Xi} \bigg( \rho(\pi,\xi) - \rho(\pi_B,\xi) \bigg)
	\]
	Here, $\Xi$ is the model uncertainty set.
	
	\cite{Petrik2016a} proposes a model based approach built on top of Robust MDPs to address this problem. They argue that the general technique of computing the robust policy and then replacing the baseline policy only if the robust policy outperform baseline is overly conservative. They show that optimizing the robust baseline regret with randomized policies is NP-hard. They propose one approximate algorithm based on the assumption that there is no error on the transition probabilities of the baseline policy. They also propose to construct and solve an RMDP to compute a better worst-case policy and replace $\pi_B$ only if it outperforms the robust performance of $\pi_B$. 
	
	\textbf{Regret based Robust} Parameter uncertainty in an MDP is common for real problems. \emph{Parametric regret} of an uncertain MDP for a policy $\pi$ is the gap between the performance of $\pi$ and that of the optimal policy $\pi^*$: 
	
	$$\max_{\pi \in \Pi}\{\rho(\pi^*)-\rho(\pi)\}$$
	
	The goal is to minimize this maximum regret, hence deriving the term \emph{MiniMax regret}. In an MDP with known parameters, minimizing regret is equivalent to maximizing the performance, but that's not the case in an uncertain MDP. In uncertain MDPs, both regret and performance of a strategy are functions of parameter realizations and they may not coincide. 
	
	\cite{Xu2009} considers uncertain MDPs where the transition laws are known, but the reward parameters are not known in advance (this is a common case in state aggregation where states are grouped together into hyper-states to build a reduced MDP. The transition law between hyper states is known in general, but the reward is uncertain due to the internal transitions inside each hyper-state). They find a MiniMax Regret (MMR) strategy by solving:
	\[
	\pi_{\text{\tiny MMR}} = \arg \min_{\pi \in \Pi} 
	\max_{r \in \mathcal{R}} \{\rho(\pi^*, r)-\rho(\pi, r)\}
	\]
	Here, $\mathcal{R}$ is the ambiguity set with admissible reward parameters and $\Pi$ is the set of admissible policies. They propose a Mixed Integer Programming (MIP) based subgradient method and show that solving this problem is NP-Hard. They reduce the problem to Linear Programs for problems with small number of vertices in $\mathcal{R}$ and small number of candidate policies in $\Pi$, which is then solvable in polynomial time.
	
	\cite{Ahmed2017} proposes more general methods to handle both reward and transition uncertainties in uncertain MDPs. They present a Mixed Integer Linear Programming (MILP) formulation to approximate the minimax regret policy for a given set of samples. They consider different scalable variants of regrets, namely \emph{Cumulative Expected Myopic Regret (CEMR)} and \emph{One Step Regret (OSR)}. CEMR for a policy $\pi$ is the cumulation of expected “maximum” possible regret in each state for that policy: 
	$$\cemr(s,\pi^t) = \sum_{a\in\actions}\pi^t(s,a)\cdot[R^{*,t}(s) - R^t(s,a)+\gamma\sum_{s' \in \states}T^t(s,a,s')\cdot \cemr^{t+1}(s',\pi^{t+1})]$$
	Here $T^t(s,a,s')$ is the transition probability of going to state $s'$ by taking action $a$ in state $s$. OSR for a given policy $\pi$ is the minimum regret over all the policies that are obtained by making changes to the policy in at most one time step: 
	$$\osr(\pi) = \min_{\hat{\pi}}[v^0(\pi^*)-v^0(\hat{\pi})] \text{ where, } \exists \text{ } \hat{t} \text{ s.t. } \forall{s,a,t} \hat{\pi}^t(s,a)=\pi^t(s,a), \text{ if } t\neq\hat{t}$$
	They show the optimal substructure property of CEMR and propose a dynamic programming based approach to minimize CEMR. They argue that CEMR is computationally expensive for dependent uncertainties across states and sparse rewards. They show that OSR overcomes these drawbacks and also present a scalable policy iteration based algorithm to optimize OSR. The novel and much simpler OSR approach consistently outperforms other regret based approaches on different practical problems in a wide margin.
	
	\section{Conclusion} \label{sec:conclusion}
	In this paper we have presented a brief survey on probabilistic reinforcement learning. We focused on two major aspects of developing RL algorithms: i) balancing the exploration-exploitation trade off and ii) computing a robust policy with limited data. We considered the underlying optimization objective as the main factor to categorize different algorithms. We presented brief descriptions along with their advantages/disadvantages and results of different methods.

\bibliography{reazul_lib}
\end{document}